\documentclass[10pt,twocolumn,letterpaper]{article}

\usepackage[pagenumbers]{cvpr} %

\usepackage{graphicx}
\usepackage{amsmath}
\usepackage{amssymb}
\usepackage{booktabs}
\usepackage{url}
\usepackage[utf8]{inputenc} %
\usepackage[T1]{fontenc}    %
\usepackage[colorlinks=true, allcolors=blue]{hyperref}       %
\usepackage{url}            %
\usepackage{booktabs}       %
\usepackage{amsfonts}       %
\usepackage{nicefrac}       %
\usepackage{microtype}      %
\usepackage{xcolor}         %
\usepackage{makecell}
\usepackage{graphicx}		%
\usepackage{algorithmic}
\usepackage{algorithm}
\usepackage{bm}

\usepackage[square,numbers]{natbib}               
\usepackage{caption}
\usepackage{subcaption}     %
\usepackage{wrapfig}        %
\usepackage{makecell}       %
\usepackage{multirow}       %

\usepackage[symbol]{footmisc}

\usepackage[capitalize]{cleveref}
\crefname{section}{Sec.}{Secs.}
\Crefname{section}{Section}{Sections}
\Crefname{table}{Table}{Tables}
\crefname{table}{Tab.}{Tabs.}

\def\cvprPaperID{11259} %
\def\confName{CVPR}
\def\confYear{2022}

\begin{document}

\title{Multi-label Iterated Learning for Image Classification with Label Ambiguity}

\renewcommand{\thefootnote}{\fnsymbol{footnote}}
\author{\textbf{Sai Rajeswar}$^{1,2,3}$\footnotemark[1], \textbf{Pau Rodríguez}$^{1}$\footnotemark[1], \textbf{Soumye Singhal}$^{2,3}$, \textbf{David Vazquez}$^1$, \textbf{Aaron Courville}$^{2,3,4}$\\
$^1$Element AI a ServiceNow company, $^2$Montréal Institute of Learning Algorithms,\\ $^3$Université de Montréal, $^4$CIFAR Fellow\\

{\tt\small rajsai24@gmail.com, pau.rodriguez@servicenow.com}
}
\maketitle

\footnotetext[1]{Equal contribution.}
\begin{abstract}
   Transfer learning from large-scale pre-trained models has become essential for many computer vision tasks. Recent studies have shown that datasets like ImageNet are weakly labeled since images with multiple object classes present are assigned a single label. This ambiguity biases models towards a single prediction, which could result in the suppression of classes that tend to co-occur in the data. Inspired by language emergence literature, we propose multi-label iterated learning (MILe) to incorporate the inductive biases of multi-label learning from single labels using the framework of iterated learning. MILe is a simple yet effective procedure that builds a multi-label description of the image by propagating binary predictions through successive generations of teacher and student networks with a learning bottleneck. Experiments show that our approach exhibits systematic benefits on  ImageNet accuracy as well as ReaL F1 score, which indicates that MILe deals better with label ambiguity than the standard training procedure, even when fine-tuning from self-supervised weights. We also show that MILe is effective reducing label noise, achieving state-of-the-art performance on real-world large-scale noisy data such as WebVision. Furthermore, MILe improves performance in class incremental settings such as IIRC and it is robust to distribution shifts. Code: \url{https://github.com/rajeswar18/MILe}
\end{abstract}

\begin{figure}[t]
    \centering
    \includegraphics[width=1.0\columnwidth,trim={25 25 25 20},clip]{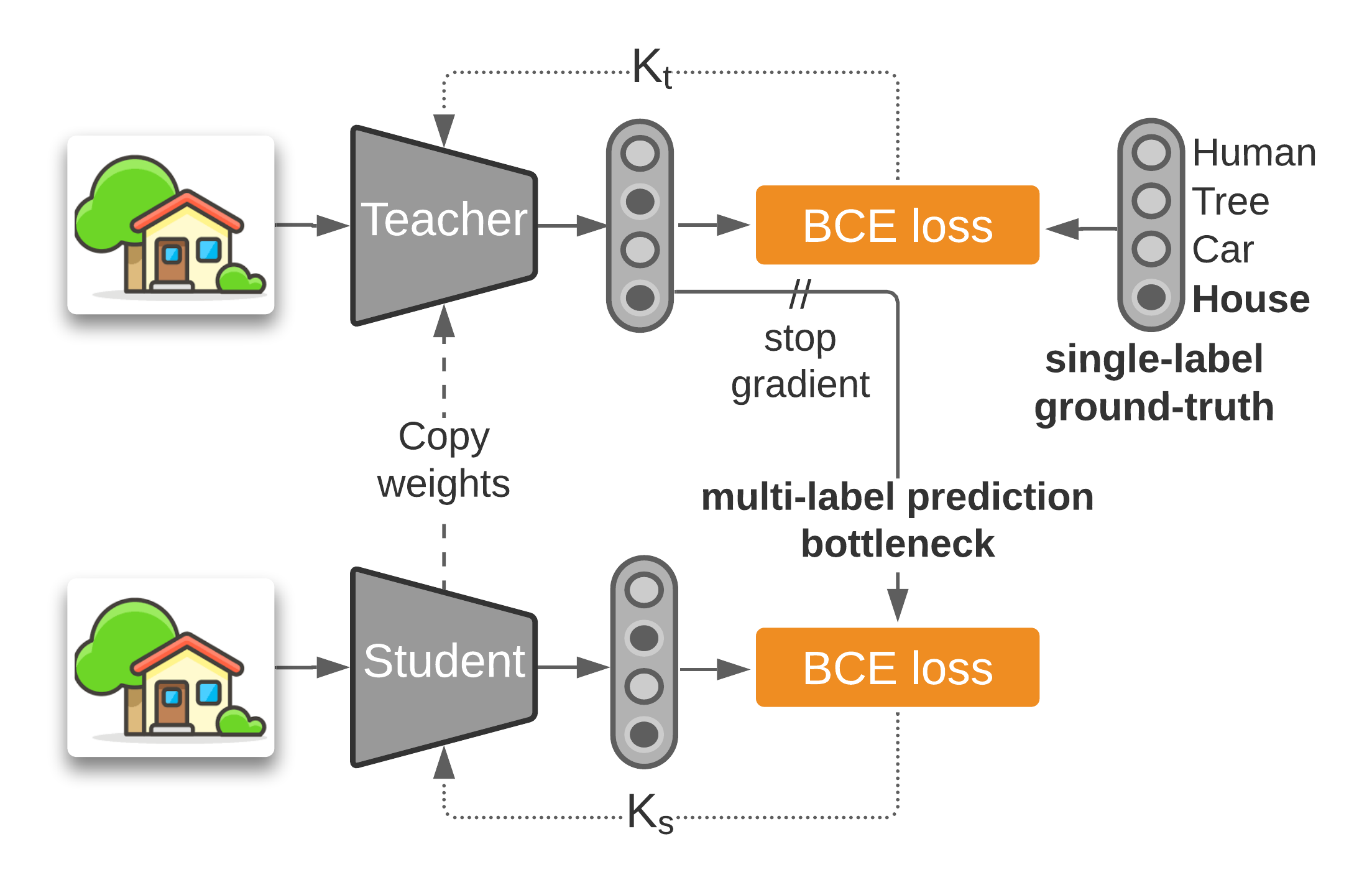}
    \caption{\textbf{Multi-label Iterated Learning (MILe)} builds a multi-label representation of the images from singly-labeled ground-truth. In this example, a model produces multi-label binary predictions for the next generation, obtaining \textit{Car} and \textit{House} for an image weakly labeled with \textit{House}.}
    \label{fig:model_overview}
\end{figure}

\section{Introduction}
Large-scale datasets with human-annotated labels have %
been central to the development of modern state-of-the-art neural network-based artificial perception systems~\citep{alexnet, he2016deep,  maskrcnn}. 
Improved performance on ImageNet~\citep{deng2009imagenet} has led to remarkable progress in tasks and domains that leverage ImageNet pretraining~\citep{QuoVA, fcnn, pspnet}.
However, these weakly-annotated datasets and models tend to project a rich, multi-label reality into a paradigm that envisions one and only one label per image. This form of simplification often hinders model performance by asking models to predict a single label, when trained on real-world images that contain multiple objects.

Given the importance of the problem, there is growing recognition of single-labeled datasets as a form of weak supervision and an increasing interest in evaluating the limits of these singly-labeled benchmarks. A series of recent studies~\citep{stock2018convnets,tsipras2020ImageNet,shankar2020evaluating,beyer2020we,yun2021re} highlight the problem of label ambiguity in ImageNet. In order to obtain a better estimate of model performance,  \citet{beyer2020we} and \citet{shankar2020evaluating} introduced multi-label evaluation sets. They identified softmax cross-entropy training as one of the main reasons for low multi-label performance since it promotes label exclusiveness. They also showed that replacing the softmax with sigmoid activations and casting the output as a set of binary classifiers results in better multi-label validation performance. 
Several other studies have explored ways to overcome the shortcomings in existing validation procedures by improving the pipelines for gathering labels~\cite{objectnet, Tsipras2020, recht19}.

In order to obtain a more complete description of images from weakly-supervised or semi-supervised data, a number of methods leverage a noisy signal such as pseudo-labels~\citep{yun2021re} or textual descriptions crawled from the web~\citep{clip}. In this work, we observe that the process of building a rich representation of data from a noisy source shares some properties with the process of language emergence studied in the cognitive science literature. In particular, \citet{Kirby2001} proposed that structured language  emerged from an inter-generational \emph{iterated learning} process~\citep{Kirby2001, Kirby2002, Kirby2014}. According to the theory, a compositional syntax emerges when agents learn by imitation from previous generations in the presence of a learning bottleneck. This bottleneck forces noisy fragments of the language to be forgotten when transmitted to new generations. Conversely, those fragments that can be reused and composed to enrich the language tend to be passed to subsequent generations. We show that the same procedure can be applied to settings that leverage a weak or noisy supervisory signal such as \citep{yun2021re,clip} to build a richer description of images while reducing the noise.  %

In this work,  
we propose multi-label iterated learning (MILe) to learn to predict rich multi-label representations from weakly supervised (single-labeled) training data.
We do so by introducing two different learning bottlenecks. First, we replace the standard convolutional neural network output softmax with a hard multi-label binary prediction. Second, we transmit these binary predictions through successive model generations, with a limited training iterations between each generation.

In our experiments, we demonstrate that MILe alleviates the label ambiguity problem by improving the F1 score of supervised and self-supervised models on the ImageNet ReaL~\citep{beyer2020we} multi-label validation set. In addition, experiments on WebVision~\citep{Li2017WebVision} show that iterated learning increases robustness to label noise and spurious correlations. Finally, we show that our approach can help in continual learning scenarios such as IIRC~\cite{abdelsalam2021iirc} where newly introduced labels co-occur with known labels. Our contributions are:

\begin{itemize}
    \item We propose MILe, a multi-label iterated learning algorithm for image classification that builds a rich multi-label representation of data from weak single labels.
    \item We show that models trained with MILe are more robust to noise and perform better on ImageNet, ImageNet-ReaL, WebVision, and multiple setups such as supervised learning (Section~\ref{sec:real}), out-of-distribution generalization (Section~\ref{sec:ood}), self-supervised fine-tuning and semi-supervised learning (Section~\ref{sec:ssl}),  and continual learning.
    \item We provide insights on the predictions made by models trained with iterated learning (Section~\ref{sec:ablations}).
\end{itemize}

\section{Related Work}

It is known that weakly-labeled datasets such as ImageNet contain label ambiguity~\citep{stock2018convnets,tsipras2020ImageNet,shankar2020evaluating,beyer2020we,yun2021re, objectnet} and label noise~\citep{van2015building, recht2019ImageNet}. Label ambiguity refers to the cases where only one of the multiple possible labels was assigned to the image. In order to evaluate how label ambiguity affects ImageNet classifiers, \citet{beyer2020we} proposed ReaL, a curated version of the ImageNet validation set with multiple labels per image. They found that ImageNet classifiers tend to perform better on ReaL since it contains less label noise but they did not address the problem of inaccurate supervision during training where more than one correct class is present in the image.
To deal with unfavorable training dynamics due to the mismatch between the multiplicity of object classes
and the majority-aggregated single labels, \citet{yun2021re}~proposed to relabel the ImageNet training set. They obtained pixel-wise labels by finetuning an ensemble of large models pretrained on a large external dataset~\cite{unreasonable17}.
Although useful, undertaking such relabeling procedure for each dataset of interest is both laborious and unrealistic. In addition, it is not clear if the same relabeling approach could be used in larger, noisier databases such as WebVision~\citep{Li2017WebVision}, which contains 2.4M images downloaded from the internet and labels consisting of the queries used to download those images. In this work, we investigate the use of iterated learning on weak singly-labeled datasets as an alternative to relabeling in order to produce a multi-label output space. Different from existing methods, MILe uses neither external data nor additional relabeling procedures. 

\paragraph{Knowledge Distillation}
Knowledge distillation is a technique commonly used in model compression~\citep{bucilua2006model,hinton2015distilling,ba2013deep}. In the vanilla setting, a large deep neural network is used as a teacher to train a smaller student network from its logits. In addition to model compression, knowledge distillation has been used to improve the generalization of student networks reusing distilled students as teachers~\cite{furlanello2018born} or distilling ensembles into a single model~\citep{allen2020towards}. Gains have been observed even when the teacher and the student model are the same network, a regime commonly known as self-distillation~\citep{mobahi2020selfdistillation,zhang2019your,allen2020towards}.
Self-distillation has also been used to improve the generalization and robustness of semi-supervised models.
\citet{xie2020self} introduced noisy student for labeling unlabeled data during semi-supervised learning. While MILe also leverages teacher and student networks, it is fundamentally different from knowledge distillation approaches. The goal of knowledge distillation is to transmit all the knowledge of a teacher network to a student network. On the other hand, MILe trains a succession of short-lived teacher and student generations, thus creating an iterated learning bottleneck~\citep{Kirby2001}, to construct a new multi-label representation of the images from single labels. This goal is also different from the goal of noisy student, which is to label unlabeled data, and which is trained three times until convergence.

\paragraph{Iterated Learning}
The iterated learning hypothesis was first proposed by \citet{Kirby2001, Kirby2002} to explain language evolution via cultural transmission in humans. Languages need to be expressive and compressible to be effectively transmitted through generations. This learning bottleneck favors languages that are compositional as they can be easily and quickly learned by the offsprings and support generalization. \citet{Kirby2014} conducted human experiments and mathematical modeling, which showed that iterated transmission of unstructured language results in convergence to a compositional language. Since then, it has seen many successful applications, especially in the emergent communication literature~\citep{guo2019emergence, ren2020compositional, cogswell2019emergence, dagan2020co}. In these settings, the learning bottleneck is induced by limiting the data or learning time of the student, which helps it to converge to a compositional language that is easier to learn~\citep{li2019ease}. The approach starts by
training a \emph{teacher network} with a small number of updates on the training set. A \emph{student network} is then trained to imitate the teacher based on pseudo-multi-labels inferred from the input samples. The student then replaces the teacher and the cycle repeats with a frequency modulated by a learning budget. Iterated learning has also been used in the preservation of linguistic structure in addition to its emergence by \citet{lu2020countering, lu2020ssil}. Furthermore, \citet{vani2021iterated} successfully applied it for emergent systematicity in VQA. To the best of our knowledge, this is the first application of the iterated learning framework in the visual domain. 

\section{Method}
\label{sec:method}

We propose MILe to counter the problem of label ambiguity in singly-labeled datasets. We delineate the details of our approach to perform multi-label classification from weak singly-labeled ground truth.

\paragraph{Enforcing Multi-label Prediction.} \label{sec:multi-label}
Singly-labeled datasets such as ImageNet usually represent their labels as one-hot vectors (all dimensions are zero except one). Training on these one-hot vectors forces models to predict a single class, even in the presence of other classes. Forcing models to predict a single class exposes them to biases in the image labeling process such as the preference for centered objects. Besides, constraining the model to output a single label per image limits the capability of perceptual models to capture all the content of the image accurately. In order to solve this problem, we propose to relax the model's output predictions from singly-labeled softmax prediction to multi-label binary prediction with sigmoids. Thus, we treat the singly-labeled classification problem as a set of independent binary classification problems. %
Since the ground-truth labels are still represented as one-hot vectors and training on them would still result in singly-labeled predictions, we propose an iterated learning procedure to bootstrap a multi-label pseudo ground truth.

\paragraph{Multi-label Iterated Learning.}
\label{sec:sil}
Our learning procedure is composed of two phases. In the first phase, a \emph{teacher} model interacts with the single-labeled data to improve its predictions. The interaction is limited to a few iterations to prevent the binary classification model from overfitting to one-hot vectors. In the second phase, we leverage the acquired knowledge to train a different model, the \emph{student}, on the multi-label predictions of the teacher. This yields a better initialization of the model for further iterations as we repeat this two-phased learning multiple times (see Alg.~\ref{alg:algorithm}).

Specifically, we consider two parametric models, the teacher $f(.; \theta_{\tau}^{T})$ and the student $f(.; \theta_{\tau}^{S})$. Parameters of the teacher $\theta_{\tau}^{T}$ are initialized using the student parameters $\theta_{\tau}^{S}$ at iteration $\tau$.  First, we train the teacher for $k_t$ learning steps on the labeled images from the dataset, obtaining $f(.;\theta_{\tau + 1}^{T})$. This constitutes the interaction phase of an iteration. We then move to the imitation phase, where we train the student to fit the teacher model for $k_s$ steps, obtaining $f(.;\theta_{\tau + 1}^{S})$. This is done by training the student on the pseudo labels generated by the teacher on the data. Finally, we instantiate a new teacher by duplicating the parameters of this new student and iterate the process until convergence. In addition to yielding a smooth transition during the imitation phase, this procedure ensures that each iteration yields an improvement over the previous one (unless it is already optimal). Note that in the supervised learning regime we do not pseudo label any unlabeled data. In Sec.~\ref{sec:ssl} we provide additional experiments showing that MILe can leverage unlabeled data in the semi-supervised learning regime.

Both the teacher and the student are trained on the same dataset $\mathcal{D}$ composed of input-label pairs $\{\mathcal{X}, \mathcal{Y}\} \in \mathcal{D}$. We train the teacher to maximize the likelihood $p(\hat{y}=y | x, \theta) = \sigma (f(x, \theta))$, where $\hat{y}$ is the label predicted by the model, $y \in \mathcal{Y}$ is the true label, and $\sigma$ is a normalization function such as the sigmoid. In order to alleviate the problem of label ambiguity, we consider $\mathcal{Y}$ a multi-label binary vector in $\mathbb{Z}_2^C$ where $C$ is the number of classes and optimize the binary cross-entropy loss:
\begin{equation}
    \mathcal{L}_{BCE} = -\frac{1}{B} \sum_{i=1}^B \sum_{j=1}^{C} y_{i, j} \cdot log(\hat{y}_{i, j}) + (1 - y_{i, j}) \cdot log(1 - \hat{y}_{i, j}), 
\end{equation}
where $B$ is the number of samples in a batch when using batched stochastic gradient descent. We show in our experiments that iterated learning along with  multi-label objective provides a strong inductive bias for modeling the effects of label ambiguity. Note that optimizing the binary cross-entropy on one-hot labels would not solve the label ambiguity problem. Thus, during each cycle, we train the teacher for a few iterations in order to prevent it from overfitting the one-hot ground truth. During student training, we threshold the teacher's output sigmoid activations to obtain multi-label pseudo ground-truth vectors $\tilde{y}=f(x,\theta^T) > \rho$. The threshold $\rho$ is $0.25$ unless otherwise stated.

\paragraph{The MILe Learning Bottleneck.}
\label{sec:learning_bottleneck}
Enforcing the imitation phase with some form of a learning budget is an essential component of the iterated learning framework~\citep{Kirby2001}. This bottleneck regularizes the student model not to be amenable to the specific irregularities in the data. \citet{Kirby2001} argue that such a  bottleneck enforces innate constraints on language acquisition. We believe that incorporating such a mechanism into the prediction models could prevent them from overfitting label noise~\citep{liu2020early}, improving the quality of pseudo labels.
There are two common ways to impose a learning bottleneck. One way is to allow a newly initialized student to only obtain the knowledge from a limited number of data instances generated by the teacher~\citep{Kirby2001, liu2021iterative}. Another is by limiting the number of student learning updates  while imitating the teacher~\citep{lu2020countering}. In our setting, we find it helpful to enforce the bottleneck via the number of learning updates. 

As illustrated in Fig.~\ref{fig:model_overview} and Alg.~\ref{alg:algorithm}, we iteratively refine a teacher network that is trained with the original labels and a student network that is trained with labels produced by the teacher. In order to prevent the student from overfitting the teacher, we restrict the amount of training updates~\citep{lu2020countering} for each of the modules. Formally, let $N$ be the size of the dataset, $k_t$ be the number of training iterations of the teacher, and $k_s$ the number of student iterations. In general, we set $k_t << N$ to prevent the teacher from overfitting one-hot labels and $k_s <= k_t$ to prevent the student from overfitting the teacher. In other words, each of our iterations is composed of two finite loops of (a) model improvement (teacher learning) and (b) model imitation (student learning).

\paragraph{Computational Cost.} \label{par:computational_cost} We train MILe for the same total number of epochs as standard supervised classification models. Thus, the total number of \textit{backward} passes through the model (counting both the teacher and the student) is the same as the standard supervised training. Thus, the only additional computational cost comes from producing pseudo-labels with the teacher model. Moreover, the pseudo-labeling only happens once per teacher-student cycle and the network is in inference mode. Assuming $k_s + k_t = 10K$ (see Figure~\ref{fig:iterationsk1k2}) and a batch size of $256$, this inference pass only happens every $2.1$ epochs for the ImageNet. Thus, the computational impact of MILe only constitutes a small fraction of the overall computational cost of training a neural network on the ImageNet. This computational cost could be easily compensated by skipping validation on alternate epochs or by validating in a different parallel process.

\begin{algorithm}[t]
  \caption{MILe}   
  \label{alg:algorithm}
    \small
    \begin{algorithmic}[1]
      \REQUIRE \textbf{Initialize} Student network $\bm{\theta}^{S}_{\tau}$, $\tau=0$.
      \hfill \COMMENT{\emph{Prepare Iterated Learning}}
      \REPEAT
      \STATE Copy $\bm{\theta}^S_\tau$ to $\bm{\theta}^{T}_{\tau + 1}$ \hfill \COMMENT{\emph{Initialize Teacher}}
      \FOR{$i=1$ {\bfseries to} $k_t$} 
      \STATE Sample a batch $(\bm{x_i}, \bm{y_i})\in \mathcal{D}_{train}$ 
      \STATE $\bm{\hat{y}_i} = f_{\bm{\theta}^{T}}(\bm{x_i}) $
      \STATE $\bm{\theta}^{T}_{\tau + 1} \leftarrow \bm{\theta}^{T}_{\tau + 1} + \alpha \nabla \mathcal{L}^{BCE}(\bm{\theta}^{T}_{\tau + 1}; \bm{y_i}, \bm{\hat{y}_i})$
      \hfill \COMMENT{\emph{Update $\bm{\theta}^T$  to minimize $L$}}
      \ENDFOR\hfill \COMMENT{\emph{Finish Interactive Learning}}
      \FOR{$i=1$ {\bfseries to} $k_s$}
      \STATE Sample a batch $(\bm{x_i}, \bm{y_i})\in \mathcal{D}_{train}$
      \STATE  $\bm{\hat{y}_i} = \sigma{(f_{\bm{\theta}^{T}_{\tau + 1}}(\bm{x_i}))} > \rho $\hfill \COMMENT{\emph{Generate Pseudo Labels}}
      \STATE $\bm{\Tilde{y}_i} = f_{\bm{\theta}^{S}}(\bm{x_i}) $
      \STATE $\bm{\theta}^{S}_{\tau} \leftarrow \bm{\theta}^{S}_{\tau} + \alpha \nabla \mathcal{L}^{BCE}(\bm{\theta}^{S}_{\tau}; \bm{\tilde{y}_i}, \bm{\hat{y}_i})$
      \hfill \COMMENT{\emph{Update $\bm{\theta}^S$  to minimize $L$}}
      \ENDFOR \hfill \COMMENT{\emph{Finish  Imitation}}
      \STATE Copy $\bm{\theta}^S_{\tau}$ to $\bm{\theta}^S_{\tau+1}$
      \STATE $\tau \leftarrow \tau + 1$
      \UNTIL{Convergence or maximum $\tau$ reached}
    \end{algorithmic}
\end{algorithm}

\begin{figure*}[!t]
    \centering
    \includegraphics[width=0.85\textwidth]{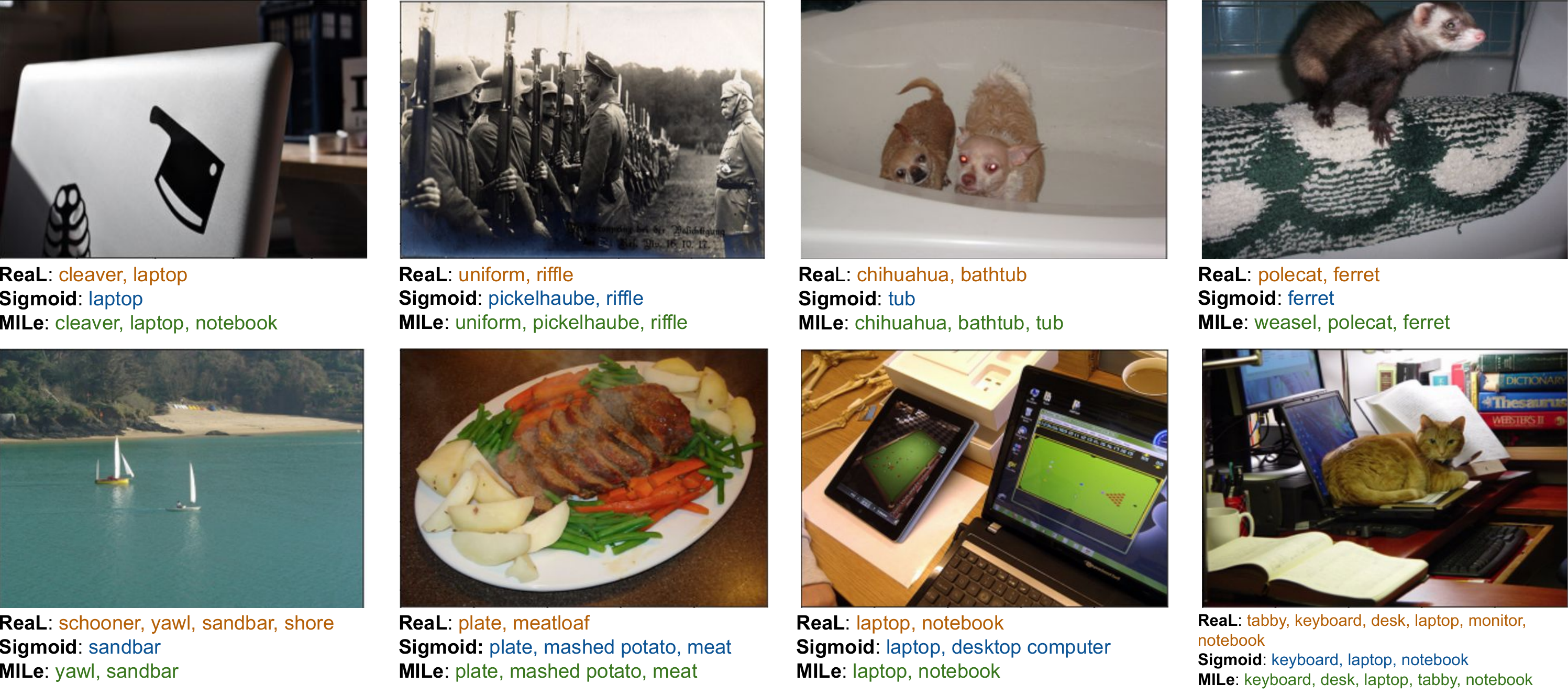}
    \caption{\textbf{Qualitative results.} ReaL: original labels. Sigmoid: ResNet-50 with sigmoid output activations. MILe: multi-label iterated learning (ours). }%
    \label{fig:qualitative_real}
\end{figure*}
\section{Experiments}
We provide experiments showing the effects of iterated learning in multiple setups. In Sec.~\ref{sec:real}, we study the robustness to label ambiguity and noise on ImageNet Real and WebVision. In Sec.~\ref{sec:ood}, we explore the benefits of iterated learning for domain generalization. In Sec.~\ref{sec:ssl}, we study the effect of MILe on models pre-trained with self-supervised objectives. Finally, in Sec.~\ref{sec:ablations}, we provide ablation experiments on the different hyperparameters as well as a more challenging synthetic setup with greater label ambiguity. Additional experiments in the Supplementary Material include a comparison with noisy student, multi-label learning on CelebA, and continual learning on IIRC.

\subsection{Label Ambiguity and Noise}
\label{sec:real}
\noindent\textbf{Datasets:} We train our models on the standard ImageNet image classification benchmark~\citep{russakovsky2015ImageNet}, which is known to contain ambiguous labels~\citep{beyer2020we}. Therefore, in addition to the validation set performance, we also report the performance on ReaL~\citep{beyer2020we}, an additional set of multi-labels for the ImageNet validation set gathered using a crowd-sourcing platform. ReaL contains a total of 57,553 labels for 46,837 images. We report results when using fractions of the total amount of training examples (i.e., 1\%, 5\%, 10\%, 100\%). To test the robustness of our method to label noise, we provide results on WebVision~\citep{Li2017WebVision}, which contains more than 2.4 million images crawled from the Flickr website and Google Images search. The same 1,000 concepts as the ImageNet ILSVRC 2012 dataset are used for querying images. It is worth noting that many ImageNet (ReaL) samples contain a single object and a single label. In Sec.~\ref{sec:multi-mnist}, we explore the limits of MILe on a synthetic dataset. 
In addition, we provide results on CelebA~\citep{liu2015faceattributes} in the supplementary material. %

\noindent\textbf{Baselines:} We train a ResNet-18 and a ResNet-50~\citep{he2016deep} model. Note that we favored vanilla ResNets over more advanced architectures and training procedures in order to focus on the advantages of MILe, rather than showing state-of-the-art results. We  compare three different methods. (i)~\emph{Softmax}: standard softmax cross-entropy loss used to train the original ResNet backbone~\citep{he2016deep}. (ii)~\emph{Sigmoid}: we substitute the cross-entropy loss for a binary cross-entropy (BCE) loss. (iii)~\emph{MILe}: the proposed method as described in Sec.~\ref{sec:method}. For WebVision experiments, we also train an additional ResNet-50-D~\citep{he2019bag} backbone following more recent methodologies~\citep{SCC}. 

\noindent\textbf{Metrics:} We report accuracy on the original~\citep{russakovsky2015ImageNet} and the ReaL~\citep{beyer2020we} ImageNet validation set. ReaL is a multi-label dataset, so we calculate the accuracy as described by~\citet{beyer2020we}. Namely, we consider a top-1 prediction correct if it coincides with any of the ground-truth labels, i.e. ReaL-Acc $= \tfrac{1}{N} \sum_{i=1}^N |\hat{y}_i \cap Y_i| > 0$, where $\hat{y}_i$ is the predicted label for the $i$th sample, $Y_i$ is the set of ReaL labels, and $|.|$ counts the the number of elements in a set. Additionally, we report the F1-score, which represents the proportion of correct predicted labels to the total number of actual and predicted labels, averaged across all examples: ReaL-F1 $= \tfrac{1}{N} \sum_{i=1}^N  \tfrac{2 \cdot TP_i}{2 \cdot TP_i + FP_i + FN_i }$, where TP is the number of true positives, FP is the number of false positives, and FN is the number of false negatives. Finally, we report the label coverage, which indicates the total fraction of labels per sample predicted by the multi-label classifier. A number 1.15 indicates an additional 15\% of labels was predicted.

\begin{table*}[!t]
\small
    \centering
\begin{tabular}{@{}c|c|cccc|cccc@{}}
\toprule
\multicolumn{1}{l|}{}           & ImageNet fraction:   & 1\%   & 5\%   & 10\%  & 100\% & 1\%   & 5\%   & 10\%  & 100\% \\ \midrule
Metric                          & Method            & \multicolumn{4}{c|}{ResNet-50} & \multicolumn{4}{c}{ResNet-18} \\ \midrule
\multirow{3}{*}{Accuracy}       & Softmax          & 6.32  & 36.71 & 53.50 & 76.33 & 6.61  & 31.5  & 48.82 & 70.41 \\
                                & Sigmoid  & 6.70  & 36.9  & 55.01 & 76.35 & 6.88  & 31.1  & 49.14 & 70.46 \\
                                & MILe (ours)              & \textbf{9.10}  & \textbf{42.52} & \textbf{57.29}  & \textbf{77.12} & \textbf{8.2}   & \textbf{36.2}  & \textbf{51.31} & \textbf{71.12} \\ \midrule
\multirow{3}{*}{ReaL-Acc}       & Softmax          & 7.19  & 42.55 & 60.21 & 82.76 & 8.80  & 35.88 & 55.11 & 77.77 \\
                                & Sigmoid  & 8.38  & 46.04 & 62.96 & 83.22 & 9.04  & 37.66 & 57.52 & 81.01 \\
                                & MILe (ours)              & \textbf{11.5}  & \textbf{48.36} & \textbf{65.42} & \textbf{83.75} & \textbf{9.18}  & \textbf{41.65} & \textbf{58.57} & \textbf{81.52} \\ \midrule
\multirow{3}{*}{ReaL-F1}        & Softmax          & 6.77  & 40.51 & 57.33 & 78.5  & 8.28  & 34.20 & 52.51 & 73.83 \\
                                & Sigmoid  & 7.17  & 41.11 & 58.46 & 78.61 & 8.39  & 33.56 & 52.12 & 73.85 \\
                                & MILe (ours)              & \textbf{10.76} & \textbf{45.02} & \textbf{62.11} & \textbf{79.89} & \textbf{8.55}  & \textbf{38.49} & \textbf{53.8}  & \textbf{74.48} \\ \midrule
\multirow{3}{*}{Label Coverage} & Softmax          & 1.00  & 1.0   & 1.0   & 1.0   & 1.0   & 1.0   & 1.0   & 1.0   \\
                                & Sigmoid  & 1.09  & 1.11  & 1.10  & 1.11  & 1.07  & 1.10  & 1.15  & 1.15  \\
                                & MILe (ours)              & 1.05  & 1.08  & 1.09  & 1.16  & 1.06  & 1.07  & 1.12  & 1.17  \\ \bottomrule
\end{tabular}
     \caption{\textbf{ImageNet results.} The first row displays the fraction of the ImageNet data used to train the models. Softmax: Vanilla ResNet with softmax loss. Sigmoid: Vanilla ResNet trained for multi-label binary classification with single labels. MILe: multi-label iterated learning. Label coverage refers to the fraction of additional labels predicted by each model. All the models are trained for 100 epochs.} %
     \label{tab:ResNet-18_ImageNet}
\end{table*}

\noindent\textbf{ImageNet results.} We report the results in Table~\ref{tab:ResNet-18_ImageNet}. MILe surpasses baseline methods on all metrics and all fractions of training data. With Sigmoid, we observe a substantial improvement on ReaL-Acc of $\sim 2\%$ and $\sim 4\%$ for ResNet-18 and ResNet-50 respectively. This is in agreement with the results reported by \citet{beyer2020we}. Incorporating iterative learning results in an extra $\sim 1\%$ performance improvement when using all the training data and up to $5\%$ of ReaL-F1 when using a smaller fraction of the data. Interestingly, we find that using smaller fractions of data reduces the label coverage.  We hypothesize that using a smaller fraction of the data leads to memorization and overfitting for the Softmax method and Sigmoid, which results in more confident predictions on a single class. Additional results focused on ReaL label recovery can be found in the supplementary material. %

We report qualitative results in Fig.~\ref{fig:qualitative_real}. As it can be seen, MILe produces more complete descriptions of the image, sometimes capturing labels that were not included in the ReaL ground truth. For instance, our method was able to detect a pickelhaube (pointy hat) that was not labeled in the ground truth. %

\begin{table}[!t]
    \setlength{\tabcolsep}{1pt}
	\small
    \centering
    \begin{tabular}{l|c|cc|cc}
     \toprule
Method & Architecture & \multicolumn{2}{c|}{WebVision} & \multicolumn{2}{c}{ImageNet} \\
 &  & Top-1&Top-5 & Top-1&Top-5 \\ \hline
   CrossEntropy~\citep{protonet}      & ResNet-50 &66.4& 83.4 &57.7 &78.4       \\                                                       
    MentorNet~\citep{mentornet}       &InceptionRes-V2& 70.8& 88.0 &62.5 &83.0        \\
    CurriculumNet~\citep{CurriculumNet}    &Inception-V2&72.1& 89.1 &64.8 &84.9            \\
    CleanNet~\citep{cleannet}      &ResNet-50&70.3& 87.8 &63.4& 84.6\\
    CurriculumNet~\citep{CurriculumNet, protonet}&ResNet-50&70.7& 88.6& 62.7 &83.4       \\
    SOM~\citep{protonet}           &ResNet-50&72.2& 89.5 &65.0& 85.1\\
    Distill~\citep{zhangdistill}   &ResNet-50&- &- &65.8 &\textbf{85.8} \\
    MoPro (dec.)~\citep{li2020mopro} & ResNet-50& 72.4 & 89.0 & 65.7 & 85.1 \\
    Multimodal~\citep{multimodal} & Inception-V3 & 73.15 & 89.73 & - & - \\\hline
    Sigmoid & ResNet-50& 72.1 & 89.5 & 65.4 & 85.0 \\
    MILe (ours)           &ResNet-50&\textbf{75.2} &\textbf{90.3} &\textbf{67.1}& 85.6\\ \bottomrule\midrule
   Initial Vanilla Model &ResNet-50-D &75.08& 89.22& 67.23 &84.09\\
SCC~\citep{SCC}& ResNet-50-D& 75.36& 89.38& 67.93 &84.77\\
SCC+GBA~\citep{SCC} & ResNet-50-D &75.69& 89.42& 68.35 &85.24\\
 \hline
    MILe (ours)           &ResNet-50-D&\textbf{76.5} &\textbf{90.9} &\textbf{68.7}& \textbf{86.4}\\
    \bottomrule
    \end{tabular}
    \caption{\textbf{WebVision results.} Methods are trained on Webvision-1000 and validated both on WebVision and ImageNet. MoPro (decoupled) is pre-trained on the same set as our method. CleanNet~\citep{cleannet} and Distill~\citep{zhangdistill} require data with clean annotations. dec: refers to "decoupled".}
    \label{tab:webvision}
\end{table}

\noindent\textbf{WebVision results.} We report results in Table~\ref{tab:webvision} and put them in context with other state of the art. For all setups, we observe that MILe attains the best performance, up to 2 points better than methods using better architectures such as Inception-V3~\citep{multimodal}. We also validate the WebVision-trained model on the ImageNet validation set, outperforming the previous state of the art and keeping results consistent with the WebVision validation set. These results suggest that the iterated learning bottleneck acts as a regularizer that prevents the model from learning noisy labels which are more difficult to fit. This hypothesis is in agreement with \citet{arpit2017closer,zhang2021understanding,liu2020early}, who showed that noise memorization happens later in the training procedure. %

\begin{table}[!t]
  \centering
    \begin{tabular}{l|c|c}
     \toprule
Method & CMNIST & CMNIST+ \\ %
\midrule
   ERM          &51.6$\pm$ 0.1 &51.1 $\pm$ 0.1  \\
    IRM~\citep{arjovsky2019invariant} &51.8$\pm$ 0.1&51.2 $\pm$ 0.2  \\
    REx~\citep{rex} &51.6$\pm$ 0.1& 51.2 $\pm$ 0.2 \\
    \hline
    MILe (ours)           &51.8$\pm$ 0.1&\textbf{53.5} $\pm$ 0.6 \\
    \bottomrule
    \end{tabular}
    \caption{\textbf{OOD generalization} on ColoredMNIST~\citep{arjovsky2019invariant} (CMNIST), which consists of predicting digits and ColoredMNIST+, which consists of color or digit prediction.}
    \label{tab:ood}
\end{table}

\subsection{Domain Generalization}\label{sec:ood}
A common problem of machine-learning models is that they tend to fail when presented with out-of-distribution data~\citep{beery2018recognition}. \citet{arjovsky2019invariant} claimed that this happens due to models relying on spurious correlations rather than the causal factors of the data. Thus, we investigate whether iterative learning can reduce the effect of spurious correlations by allowing the model to produce independent predictions of the two correlated factors. Following \citet{arjovsky2019invariant}, we perform experiments on ColoredMNIST~\citep{arjovsky2019invariant}, a version of MNIST where the color of the digits is spuriously correlated with their value. The spurious correlation is removed at test time, i.e. colors are assigned randomly, to reveal whether models are affected by color. During training, we add an extra color classification task consisting of solid color images. For each task, models are either asked to predict the color or the digit but never both. This setup brings ColoredMNIST closer to ImageNet's label ambiguity problem, where labels are biased towards foreground (e.g., a cow on a beach) but backgrounds (beaches) are also part of the classification problem. We call this setup ColoredMNIST+ (details in the supplementary material %

\noindent\textbf{Results.} We compare with invariant risk minimization (IRM)~\citep{arjovsky2019invariant} and risk extrapolation (REx)~\citep{rex} based on the DomainBed implementation~\citep{gulrajani2020search}. These two approaches leverage differences between multiple environments, with different levels of correlation between digit and color, to become invariant to spurious attributes.%
We report results in Table~\ref{tab:ood}. MILe surpasses REx by 2 points. Interestingly, even though ERM and IRM are also required to solve the color classification task, only iterated learning is able to use it to improve performance. Although the color and digit prediction tasks are mutually exclusive, during iterated learning the teacher produces labels for both tasks simultaneously and thus the student learns to predict the color even for images that contain a digit. This helps the model to learn that these are two independent attributes, boosting its performance.

\begin{table}[!t]
    \small
    \centering

\setlength{\tabcolsep}{1pt}
\begin{tabular}{@{}l|ccc|ccc@{}}
\toprule
\multicolumn{1}{c|}{Method}  & \multicolumn{3}{c|}{ImageNet Validation}                              & \multicolumn{3}{c}{ImageNet ReaL-F1}             \\
                     & 1\%   & 10\% & \multicolumn{1}{c|}{100\%} & 1\%   & 10\%  & 100\% \\ \midrule
SimCLR~\citep{chen2020simple}               & 48.3  & 65.6 & \multicolumn{1}{c|}{76.25} & 51.54 & 69.16 & 76.91 \\
BYOL~\citep{grill2020bootstrap}                 & 53.2  & 68.8 & \multicolumn{1}{c|}{77.2}  & 54.32 & 70.81 & 78.85 \\
SwAV~\citep{caron2020unsupervised}                 & 53.9  & 70.2 & \multicolumn{1}{c|}{77.74} & 55.79 & 71.22 & 79.18 \\ \midrule
MoCo-v2~\citep{chen2020improved}               & 51.72 & 66.5 & \multicolumn{1}{c|}{77.12} & 53.34 & 70.75 & 79.04 \\
MILe (Ours) + \citep{chen2020improved}        & \textbf{52.62} & \textbf{67.4} & \multicolumn{1}{c|}{\textbf{77.38}} & \textbf{56.08} & \textbf{71.48} & \textbf{80.03} \\ \midrule
SimCLR-v2-sk0~\citep{chen2020big}        & 58.18 & 68.9 & \multicolumn{1}{c|}{76.3}  & 57.25 & 70.11 & 78.83 \\
MILe (Ours) + \citep{chen2020big} (sk0) & \textbf{61.85} & \textbf{70.5} & \multicolumn{1}{c|}{\textbf{77.29}} & \textbf{60.49} & \textbf{72.76} & \textbf{79.38} \\
SimCLR-v2-sk1~\citep{chen2020big}        & 64.7  & 72.4 & \multicolumn{1}{c|}{78.7}  & 62.77 & 74.21 & 79.43 \\
MILe (Ours) + \citep{chen2020big} (sk1) & \textbf{69.4}  & \textbf{74.7} & \multicolumn{1}{c|}{\textbf{79.5}}  & \textbf{65.04} & \textbf{76.40} & \textbf{81.53} \\ \bottomrule
\end{tabular}

     \caption{\textbf{Self-supervised finetuning.} The second row displays the fraction of ImageNet training data used for fine-tuning. Accuracy of top-1 predictions are used for reporting the numbers.}
    
     \label{tab:ssl}
\end{table}

    \begin{table}[t]
    \centering
    \begin{tabular}{c|c|cc}
        \toprule
      \makecell{Method}  & Teacher & \multicolumn{2}{c}{\makecell{Label fraction}} \\
      && 1\%&10\% \\ \midrule
                 Distilled~\citep{chen2020big} & R50 (2×+SK)&69.0&75.1\\
                  \hline
                 Self-distilled~\citep{chen2020big} & R50 (1x+SK)   &70.15  &  74.43  \\
                 MILe (ours) & R50 (1x+SK)& \textbf{73.08} & \textbf{75.3}\\
        \bottomrule
       
    \end{tabular}
     \caption{\textbf{Self- semi-supervised learning.} ImageNet top-1 accuracy for ResNet-50 (R50) distilled from a SimCLR~\citep{chen2020simple} model. 2$\times$: teacher has $2 \times$ parameters than the student. }
    
     \label{tab:semi_supervised}
\end{table}

\subsection{Self-supervised Fine-tuning}
\label{sec:ssl}
ImageNet's label ambiguity~\citep{stock2018convnets,tsipras2020ImageNet,shankar2020evaluating,beyer2020we,yun2021re} might be problematic for fully-supervised methods but it is possible that self-supervised pre-training procedures such as MoCo~\citep{he2019moco} or SimCLR~\citep{chen2020simple} are immune to it. We explore whether iterated learning improves the performance of self-supervised models in the fully- and semi-supervised fine-tuning regimes. We perform experiments on the ImageNet dataset and report validation accuracy and ReaL-F1 as described in Sec.~\ref{sec:real}.

\noindent\textbf{Baselines.} We report results with ResNet-50 pre-trained with  SimCLR~\citep{chen2020simple}, SimCLR-v2~\citep{chen2020big}, BYOL~\citep{grill2020bootstrap}, MoCo-v2~\citep{chen2020improved}, and SwAV~\citep{goyal2021self}. Results are reported after fine-tuning weights with 1\%, 10\%, and 100\% of the ImageNet training set. We incorporate the proposed iterative learning procedure in the fine-tuning process of MoCo-v2 and SimCLR-v2. For SimCLR-v2, we also tested the "sk1" variant which was improved with selective kernels~\citep{li2019selective,chen2020big}, while "sk0" is the vanilla version. For the semi-supervised learning experiments, we compare with SimCLR-v2's distillation experiments, where a teacher predicts pseudo-labels on unlabeled data. We compare with ResNet-50 (2$\times$+SK), where the teacher has $2\times$ capacity than the student, and ResNet-50 (1$\times$+SK) where the teacher and the student are the same models.

\begin{figure}[t]
    \centering
    \begin{subfigure}[t]{0.49\linewidth}
    \includegraphics[width=\linewidth,trim={00 00 10 10},clip]{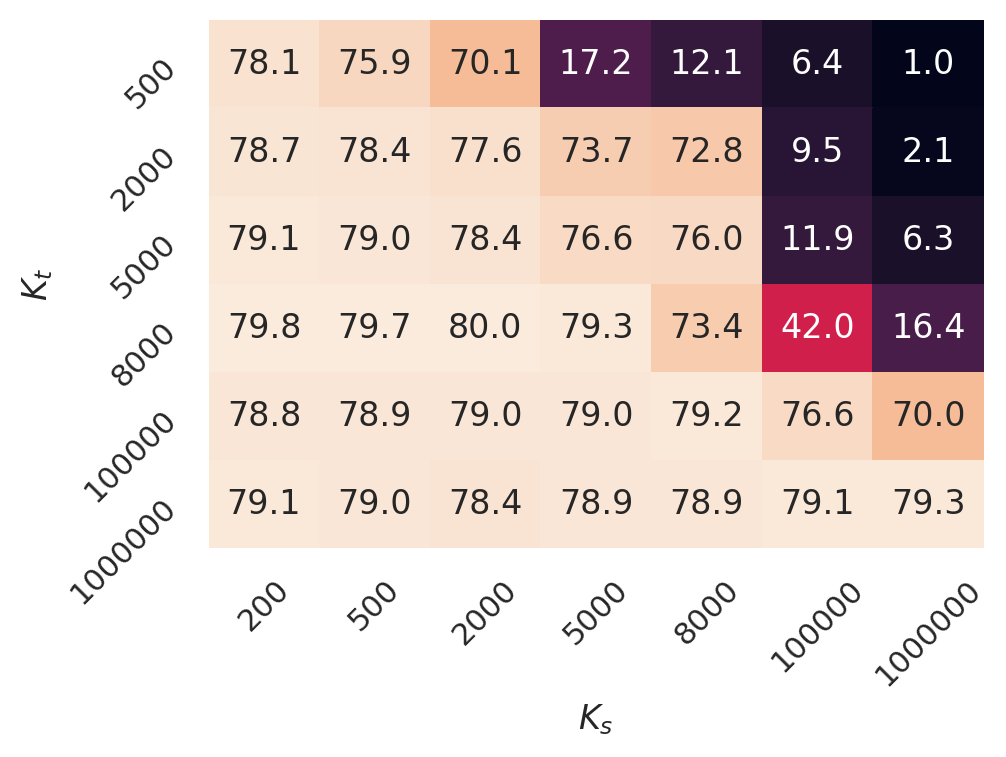}
    \caption{Iterations}
    \label{fig:iterations_abaltion}
    \end{subfigure}\hfill
    \begin{subfigure}[t]{0.49\linewidth}
    \includegraphics[width=\linewidth,trim={0 -40 20 20},clip]{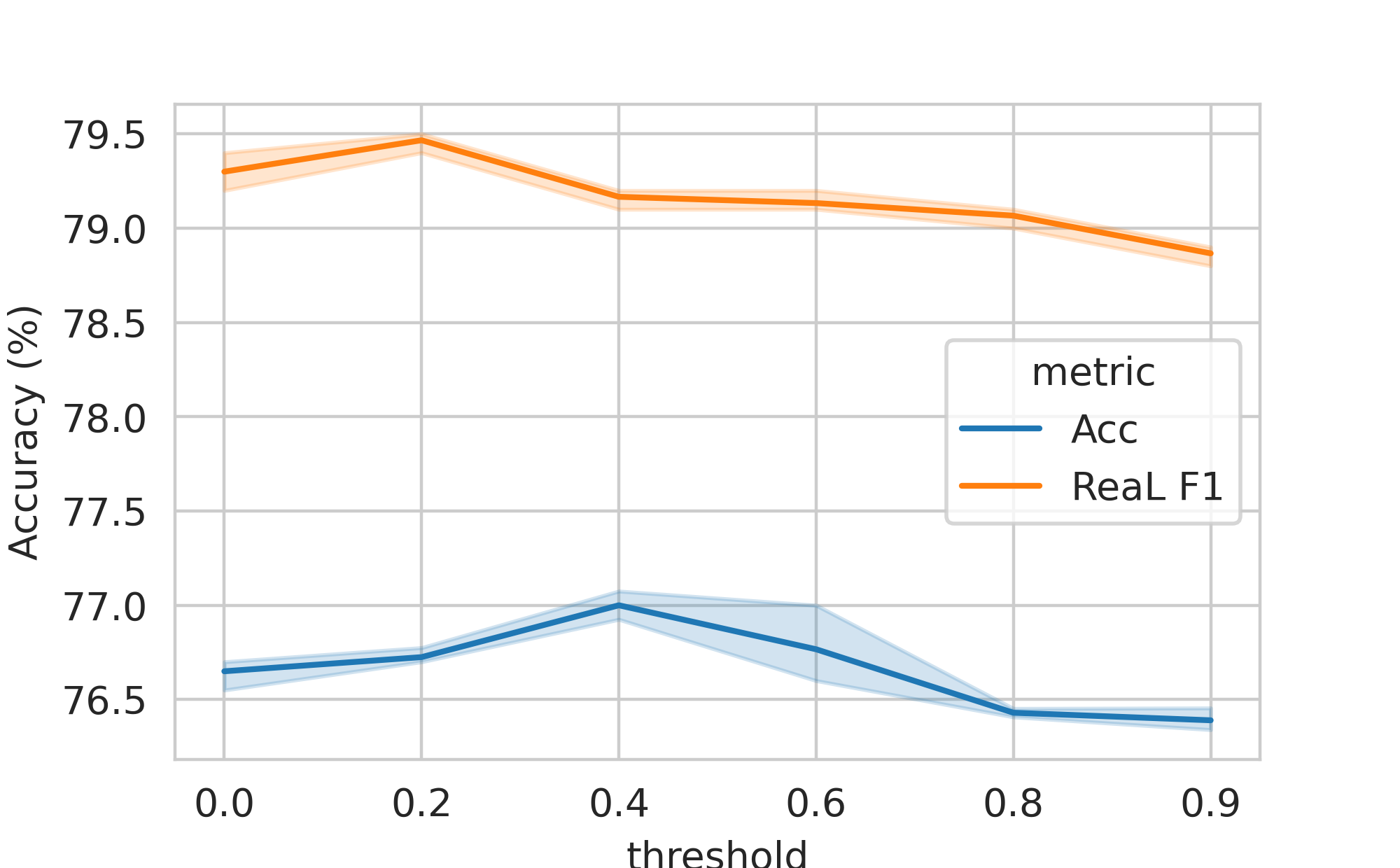}
    \caption{Threshold}
    \label{fig:threshold_ablation}
    \end{subfigure}
    
    \caption{\textbf{Ablation study.} Comparison between different iteration schedules. (a) Sweep over length of teacher training $k_t$ and length of student training $k_s$. We report the ReaL-F1 score. (b)  ReaL F-1 and accuracy scores for a threshold value sweep ($\rho$).}
    \label{fig:iterationsk1k2}
\end{figure}
\noindent\textbf{Results.} We report fine-tuning results in Table~\ref{tab:ssl}. Iterated learning improves the performance of MoCo-v2, SimCLR, and SimCLR-v2 for all fine-tuning data fractions. Interestingly, the improvement gap grows when using better self-supervised initializations. For example, the ReaL improvement from the best performing SimCLR-v2-sk1 with 100\% of the validation data is $4.6\%$ while it is around $3\%$ for MoCo-v2 and SimCLR-v2-sk0. We hypothesize that more accurate models lead to better teachers, improving the overall performance of the iterated learning procedure. 

We report semi-supervised learning results in Table~\ref{tab:semi_supervised}. Iterated learning performs 2.9\% better with 1\% of the training labels and 0.9\% with 10\% of the training labels when compared with the self-distillation procedure presented in SimCLR-v2~\cite{chen2020big}. Interestingly, we find that iterated learning attains better performance than distilling from a teacher twice the size of the student.

\subsection{Analysis}
\label{sec:ablations}
In this section we explore the behavior of MILe under different hyperparameter settings as well as more challenging setups with synthetic data. 

\paragraph{Number of Iterations.}
We investigate the effect of the number of teacher iterations ($k_t$) and student iterations ($k_s$) per cycle on the final performance (Fig. \ref{fig:iterations_abaltion}). We report the ReaL-F1 for different $k_t$ values (rows) and $k_s$ values (columns). In general, we find that good performance can be achieved with a wide range of $k_t$ and $k_s$ combinations. The best performance is achieved with smaller values of $k_t$ and $k_s$. Extreme values of $k_t$ and $k_s$ lead to lower performance, with the model being most sensitive to large values of $k_s$ (dark regions). This is expected since a small $k_t$ would let the imitation phase constantly disrupt supervised learning via interaction with the data, while a large $k_t$ does not reap the benefits of distillation. For a given $k_t$ we find that the optimal $k_s$ lies in the mid-range and the other way around. Regarding the influence of the dataset size, we observe that it mostly influences the optimal number of teacher iterations ($k_t$). We hypothesize that it takes few iterations for the teacher to overfit small datasets, which leads to one-hot predictions and prevents the model from learning a multi-label hierarchy.

\paragraph{Pseudo-label Threshold Ablation Study}
\label{app:threshold}
In this section, we conduct an ablation study on the threshold value ($\rho$) used by MILe to produce multi-pseudo-labels from sigmoid output activations (see Section~\ref{sec:method} and Algorithm~\ref{alg:algorithm}).  Fig.~\ref{fig:threshold_ablation} shows the validation accuracies and  ReaL-F1 scores for different threshold values. Lower thresholds bias the student towards producing multi-label outputs, even for low-confidence classes. Larger threshold values make the student tend towards singly-labeled prediction, only predicting labels for which the confidence is high. In the extreme, a high threshold constrains the teacher to predict empty label vectors. Interestingly, we find that lower threshold values result in higher ReaL-F1 score and better accuracy. In fact, the Real-F1 score benefits from lower $\rho$ than the accuracy. This is due to the fact that lower thresholds increase the number of predicted labels per image, which improves the recall in multi-label evaluations.

\paragraph{Multi-label MNIST}
\label{sec:multi-mnist}

\begin{table}[t!]
\centering
\begin{tabular}{@{}lcc@{}}
\toprule
        & \textbf{F1@0.25} & \textbf{F1@0.5} \\ \midrule
Softmax & 28.69            & 28.69           \\
Sigmoid & 29.10            & 28.67        \\
MILe (ours)    & \textbf{41.35}   & \textbf{34.32}  \\ \bottomrule
\end{tabular}
\renewcommand{\figurename}{Table}
\caption{\textbf{Results on multi-label MNIST.} The first column displays the F1 score when the threshold for positive labels is set to 0.25 and the second column shows the F1 score for a threshold of 0.5.}
\label{tab:multi-mnist}
\renewcommand{\figurename}{Figure}
\end{table}
\begin{figure}[t!]
\centering
\includegraphics[width=0.8\linewidth]{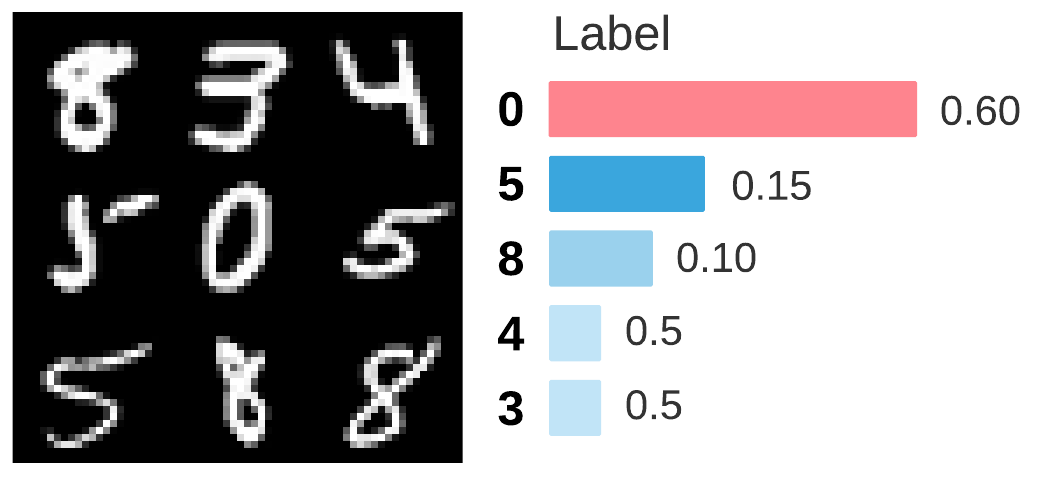}
\caption{\textbf{Multi-MNIST.} The center digit has a probability of $0.6$ to be chosen as the label for the whole grid. }%
\label{fig:multi-mnist}
\end{figure}

Many images in the real world datasets like WebVision or ImageNet contain a single object, which biases MILe towards predicting a small number of objects per image. In order to explore the limits of MILe, we begin by designing a controlled experiment on a synthetic dataset where most samples contain multiple classes. Each sample consists of a $3\times3$ grid of randomly sampled MNIST digits~\citep{lecun-mnisthandwrittendigit-2010}. For each grid, its single label corresponds to the center digit with probability $0.6$ while the $8$ remaining digits are sampled with probability $0.05$ each (see Fig.~\ref{fig:multi-mnist}). Note that, similar to the ImageNet, digits of the same class can repeat in the grid. However, the probability of having a $3\times3$ grid with the same digit repeated in each position is $10^{-9}$.

Results are shown in Table~\ref{tab:multi-mnist}. We observe that MILe attains up to 12\% better F1 score than the Softmax and Sigmoid baselines. It is worth noting that the improvement is most significant when thresholding the sigmoid output predictions to $0.25$. Interestingly, for this experiment, we found the best threshold to produce multi-pseudo-labels from the teacher output to be ($\rho=0.1$). Having a low threshold biases the student towards producing multi-label outputs. We find these results encouraging and we believe that better performance could be attained by improving the pseudo-multi-label generation strategy. We plan to explore these new strategies in future work.

\section{Discussion}
We introduce multi-label iterated learning (MILe) to address the problem of label ambiguity and label noise in popular classification datasets such as ImageNet. MILe leverages iterated learning to build a rich supervisory signal from weak supervision. It relaxes the singly-labeled classification problem to multi-label binary classification and alternates the training of a teacher and a student network to build a multi-label description of an image from single labels. The teacher and the student are trained for few iterations in order to prevent them from overfitting the singly-labeled noisy predictions. MILe improves the performance of image classifiers for the singly-labeled and multi-label problems, domain generalization, semi-supervised learning, and continual learning on IIRC. A possible limitation, which is inherent to iterated learning~\citep{lu2020countering}, is choosing the correct length of teacher ($k_t$) and student iterations ($k_s$). However, our ablation experiments suggest that the proposed procedure is beneficial for a wide range of $k_t$ and $k_s$ values (Sec.~\ref{sec:ablations}). MILe also depends on the threshold value $\rho$, which we use to produce pseudo-labels from the teacher's outputs. However, we found encouraging that low values of $\rho$ improve the performance of the classifiers, indicating that predicting multiple labels is beneficial. With respect to the computational cost, we found that the impact of MILe is lower than the validation phase of the models (see Sec.~\ref{par:computational_cost}).  
Overall, we found that iterated learning improves the performance of models trained with weakly labeled data, helping them to overcome problems related to label ambiguity and noise. We hope that our research will open new avenues for iterated learning in the visual domain.

{\small
\bibliographystyle{abbrvnat}
\bibliography{egbib}
}
\clearpage
\appendix
\section*{Supplementary Material}
 In Section~\ref{sec:iirc}, we provide continual learning results on the IRCC benchmark~\citep{abdelsalam2021iirc}. In Section~\ref{app:background} we investigate to which extent MILe is able to recover labels that were not present in the original dataset. In Section~\ref{app:domainbed} we provide additional details on the domain generalization experiment. In Section~\ref{app:celeba}, we provide additional results for multi-label classification on CelebA. In Section~\ref{app:noisy_student}, we test additional iterated learning schedules such as that of noisy student.

\section{IIRC benchmark}
\label{sec:iirc}
We explore whether MILe can incrementally learn an increasingly complex class hierarchy by teaching previously seen tasks to new generations. We experiment with Incremental Implicitly-Refined Classification (IIRC)~\citep{abdelsalam2021iirc}, an extension to the class incremental learning setup~\citep{masana2020class} where the incoming batches of classes have two granularity levels, e.g. a coarse and a fine label. Labels are seen one at a time, and fine labels for a given coarse class are introduced after that coarser class is visited. The goal is to incorporate new finer-grained information into existing knowledge in a similar way as humans learn different breeds of dogs after learning the concept of dog.

\subsection{Metrics}
\label{app:jaccard}

\begin{figure}[t]
    \begin{subfigure}{\linewidth}
\centering
    \includegraphics[width=0.49\linewidth]{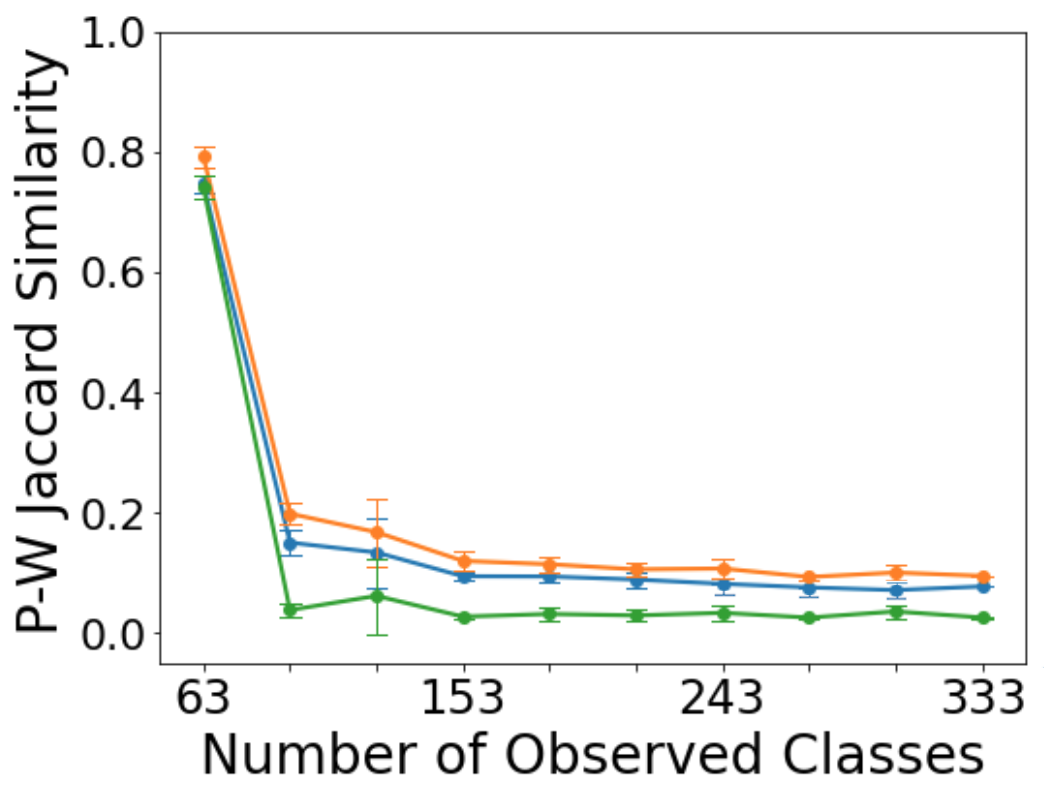}
    \includegraphics[width=0.49\linewidth]{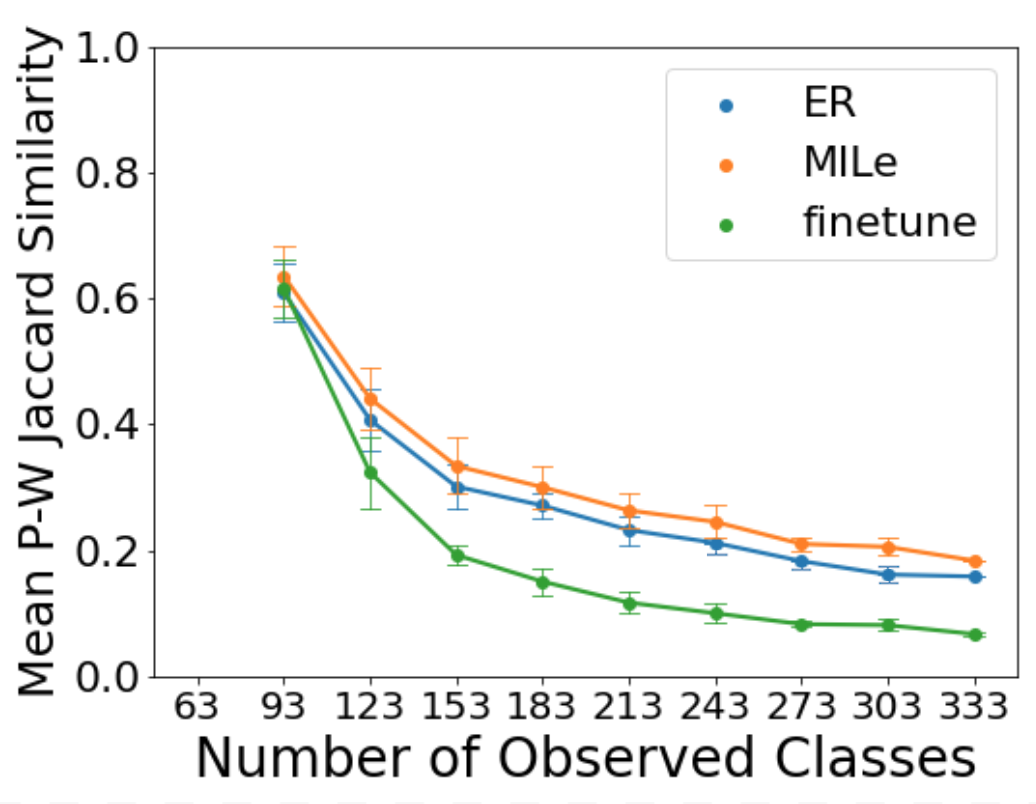}
    \caption{IIRC-ImageNet}
\label{fig:iirc_imagenet}
\end{subfigure}
\begin{subfigure}{\linewidth}
\centering
    \includegraphics[width=0.49\linewidth]{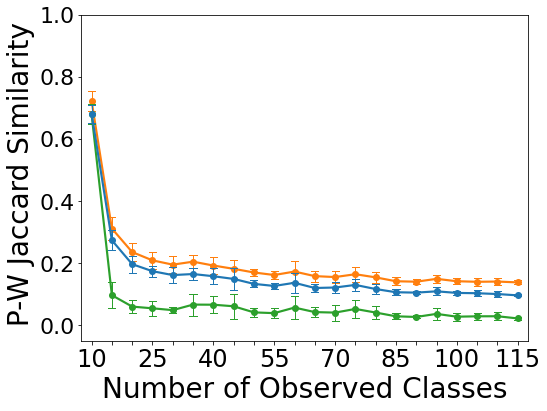}
    \includegraphics[width=0.49\linewidth]{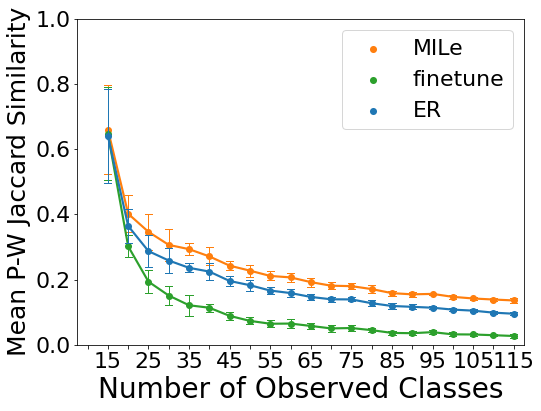}
     \caption{IIRC-CIFAR10}
\label{fig:iirc_CIFAR}
\end{subfigure}
    \caption{\textbf{IIRC evaluation}. (a) Average performance on IIRC-ImageNet-lite. (b) Average performance on IIRC-CIFAR10. We run experiments on five different task configurations and report the mean and standard deviation. 
    \label{fig:iirc}
    Left: average performance when the tasks are equally weighted irrespective of how many samples exist per task.
Right: average performance over the number of samples. In this case, the first task has more weight since it is larger in the number of samples.}
    \label{fig:iirc_imgenetlite}
\end{figure}

 As it can be seen in Fig.~\ref{fig:iirc}, the two reported metrics are the precision-weighted Jaccard similarity and the mean precision-weighted Jaccard similarity.

\paragraph{Precision-weighted Jaccard Similarity.} The Jaccard similarity (JS) refers to the intersection over union between model predictions $\hat{Y}_i$ and ground truth $Y_i$ for the $i$th sample:
\begin{equation}
JS = \frac{1}{n}\sum_{i=1}^n \frac{|Y_i\cap \hat{Y_i}|}{|Y_i\cup  \hat{Y_i}|},
\end{equation}
The precision-weighted JS for task $k$ is the product between the JS and the precision for the samples belonging to that task:
$$R_{jk} = \frac{1}{n_k}\sum_{i=1}^{n_k} \frac{|Y_{ik}\cap \hat{Y}_{ik}|}{|Y_{ik}\cup  \hat{Y}_{ik}|} \times \frac{|Y_{ik}\cap \hat{Y}_{ik}|}{\hat{Y}_{ik}}$$

where $(j \geq k), \hat{Y}_{ik}$ is the set of (model) predictions for the $i$th sample in the $k$th task, $Y_{ik}$ are the ground truth labels, and $n_k$ is number of samples in the task. $R_{jk}$ can be used as a proxy for the model’s performance on the $k$th task as it trains on more tasks (i.e. as j increases).

\paragraph{Mean precision-weighted Jaccard similarity.} We evaluate the overall performance of the model after training until the task $j$, as the average precision-weighted Jaccard similarity over all the classes that the model has encountered so far. Note that during this evaluation, the model has to predict all the correct labels for a given sample, even if the labels were seen across different tasks.

\subsection{Results.} Following the procedure described by~\citet{abdelsalam2021iirc}, we train a ResNet-50 on ImageNet and a reduced ResNet-32 on CIFAR100. Also following~\citet{abdelsalam2021iirc}, we compare with an \textit{experience replay} (ER) baseline and a \textit{finetune} lower-bound. %
We report the model's overall performance after training until task $i$ as the precision-weighted Jaccard similarity between the model predictions and the ground-truth multi-labels over all classes encountered so far.
We report IIRC-ImageNet-lite evaluation scores in Fig.~\ref{fig:iirc_imagenet} and CIFAR in Fig.~\ref{fig:iirc_CIFAR}. In all cases, we find that iterative learning increases the performance with respect to the ER baseline by a constant factor. This suggests that MILe helps prevent forgetting previously seen labels by propagating them through the iterated learning procedure.

\section{ReaL label recovery}
\label{app:background}
The goal of MILe is to alleviate the problem of label ambiguity by recovering all the alternative labels for a given sample. We define alternative labels as those that were not originally present in the ground truth. In this section, we evaluate how much of those alternative labels are recovered with MILe.

\begin{table}[h]
	\small
    \centering
    \begin{tabular}{l|ll|ll}
     \toprule
Method &  \multicolumn{2}{c}{ResNet-50} & \multicolumn{2}{c}{ResNet-18} \\
 &   10\% data &100\% data & 10\% data&100\% data \\ \hline
    Softmax &0.2171 &0.2679 &0.1983 & 0.2648\\
    Sigmoid &0.2310 &0.2845 &0.2047 & 0.2836\\
    \hline
    MILe (ours)           &\textbf{0.3042}&\textbf{0.3248} &\textbf{0.2187} &\textbf{0.2880}\\
     \bottomrule
    \end{tabular}
    \caption{\small \textbf{Secondary label recovery.} Mean average precision over labels that appear in ReaL but not in the original ImageNet validation set. }
    \label{tab:recovery}
\end{table}

Table~\ref{tab:recovery} displays the mean average precision on the alternative labels present in ReaL~\citep{beyer2020we}. As it can be seen, MILe is able to recover up to 7\% more labels than replacing softmax by sigmoid and binary cross entropy during training.

\section{Details on Domain Generalization}
\label{app:domainbed}
\begin{figure}[h]
\centering
\includegraphics[width=\linewidth]{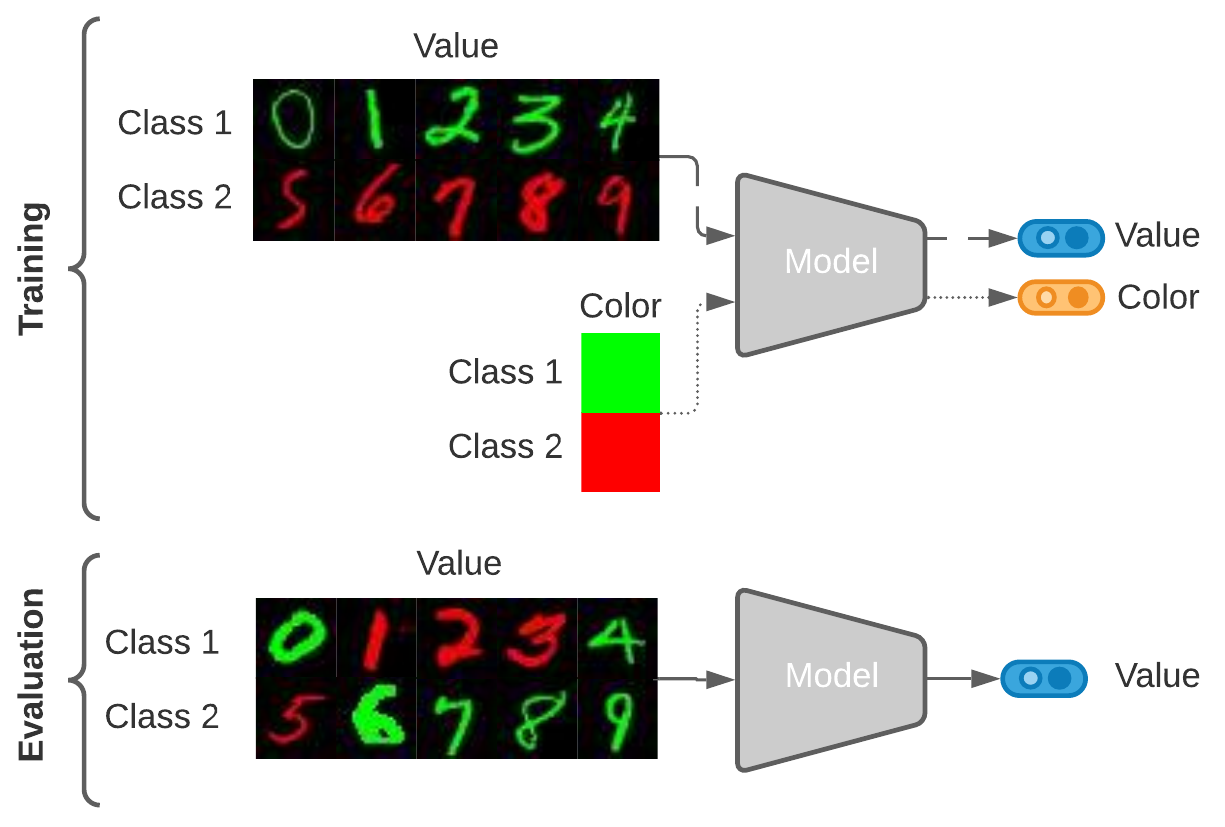}
\caption{\textbf{ColoredMNIST+.} During training, the model is asked to classifier either digits or colors. Digits are highly correlated with their color, e.g. 0-4 tend to be green while 5-9 tend to be red. At test time, digits are less correlated with color.}
\label{fig:mnistplus}
\end{figure}
In order to investigate how models perform outside of their original training distribution, \citet{arjovsky2019invariant} introduced ColoredMNIST, a dataset of digits presented in different colors. In order to create spurious correlations, the color of the digits is highly correlated with the value itself. During training, data is sampled from two different image-label distributions or environments. In the first one, the correlation between digit and color is 90\% and in the second is 80\%. The correlation between the digit and color is 10\% at test time. Since we want to explore the effect on generalization when the model is able to predict the digit and the color independently, we add a 33\% chance of showing a blank image with no digit and only background color, where the background color is the label. This would be equivalent to a "beach" class in ImageNet. Note that this change does not remove the spurious correlations between the existing digits and their color. We call this benchmark ColoredMNIST+, see Fig.~\ref{fig:mnistplus}. During training, iterated learning builds a multi-label represenation of the digits, often including their color, leading to better disentanglement of the concepts "digits" and "color".

\section{Multi-label classification on CelebA}
\label{app:celeba}
We provide results on CelebA~\citep{liu2015faceattributes}, a multi-label dataset. CelebA is a large-scale dataset of facial attributes with more than 200K celebrity images, each with 40 attribute annotations that are known to be noisy~\citep{speth2019automated}.
We report results in Table~\ref{tab:celeba}. Interestingly, despite the fact that CelebA is a multi-label dataset, we observe a $\sim 1\%$ improvement in F1 score when using the proposed iterative learning procedure. This along with per-class balanced accuracy in Table~\ref{table:att_acc_pred} is in line with our hypothesis that the iterated learning bottleneck has a regularization effect that prevents the model from learning noisy labels~\citep{lu2020countering}. It is worth noting that MILe shows improved scores for the attributes that are difficult to classify such as  \emph{big-lips}, \emph{arched-eyebrows} and \emph{moustache}.

\begin{table}[t!]
	\small
    \centering
    \begin{tabular}{l|l}
     \toprule
Method &  
  F1-score\\ \hline
   
    CE-Sigmoid &80.14  \\
    ResNet-18(FPR)~\citep{bekele} & 77.55\\
    ResNet-34 (FPR)~\citep{bekele} &79.96 \\
    \hline
    MILe (ours)           &\textbf{81.40} \\
     \bottomrule
    \end{tabular}
    \caption{\small Comparison on CelebA multi-attribute classification. Just as in ReaL ImageNet validation, we use F1-score (based on the intersection over union) measure to evaluate the methods. }
    \label{tab:celeba}
\end{table}

\begin{table*}[t!]
  \small
  \centering
  \newcommand*\rot{\rotatebox{80}}

  \scalebox{0.62}{
    \begin{tabular}{@{} cl*{20}c @{}}
      \toprule
      & & \rot{5 o Clock Shadow } &\rot{Arched Eyebrows} &  \rot{  Attractive } & \rot{  Bags Under Eyes} & \rot{  Bald } &\rot{  Bangs} & \rot{  Big Lips} &    \rot{  Big Nose} &\rot{  Black Hair} & \rot{  Blond Hair} &  \rot{  Blurry  } &   \rot{  Brown Hair } & \rot{  Bushy Eyebrows       } &  \rot{  Chubby } & \rot{  Double Chin          } &  \rot{  Eyeglasses } &  \rot{  Goatee  } & \rot{  Gray Hair            } &            \rot{  Heavy Makeup         } &  \rot{  High Cheekbones      } \\
      & Triplet-kNN~\cite{schroff2015facenet} &  66          & 73          & 83          & 63          & 75          & 81          & 55          & 68          & 82          & 81          & 43          & 76          & 68          & 64          & 60          & 82          & 73          & 72          & 88          & 86           \\ 
      & PANDA~\cite{zhang2014panda} &  76    & 77    & 85    & 67    & 74    & 92    & 56    & 72    & 84    & 91    & 50    & \textbf{85}    & 74    & 65    & 64    & 88    & 84    & 79    & 95    & \textbf{89}     \\ 
      & Anet~\cite{liu2015faceattributes}       &  81          & 76          & \textbf{87}          & 70          & 73          & 90          & 57          & \textbf{78}          & \textbf{90}          & 90          & 56          & 83          & 82 & 70          & 68          & 95          & \textbf{86}          & 85          & \textbf{96}          & \textbf{89}           \\ 

      & MILe & \textbf{85} & \textbf{83} & 82& \textbf{74}& \textbf{82} & \textbf{92}&\textbf{65}&74&88&\textbf{91}&\textbf{76}&79&\textbf{83}&\textbf{72}&\textbf{72}&\textbf{98}&\textbf{86}&\textbf{86}&93&\textbf{89}\\
      \midrule
      & & \rot{  Male } &\rot{  Mouth Slightly Open } &    \rot{  Mustache} & \rot{  Narrow Eyes} &  \rot{  No Beard } & \rot{  Oval Face } &  \rot{  Pale Skin} & \rot{  Pointy Nose} & \rot{  Receding Hairline    } & \rot{  Rosy Cheeks} & \rot{  Sideburns } & \rot{  Smiling } &\rot{  Straight Hair } & \rot{  Wavy Hair } &  \rot{  Wearing Earrings} &  \rot{  Wearing Hat  } & \rot{  Wearing Lipstick } & \rot{  Wearing Necklace} &  \rot{  Wearing Necktie} & \rot{  Young  } \\

      & Triplet-kNN~\cite{schroff2015facenet} &  91          & 92          & 57          & 47          & 82          & 61          & 63          & 61          & 60          & 64          & 71          & 92          & 63          & 77          & 69          & 84          & 91          & 50          & 73          & 75           \\ 
      & PANDA~\cite{zhang2014panda} &  \textbf{99}    & 93    & 63    & 51    & 87    & 66    & 69    & 67    & 67    & 68    & 81    & \textbf{98}    & 66    & 78    & 77    & 90    & \textbf{97}    & 51    & \textbf{85}    & 78     \\ 
      & Anet~\cite{liu2015faceattributes}        &  \textbf{99}          & \textbf{96}          & 61          & 57          & 93          & \textbf{67}          & \textbf{77}          & 69          & 70          & \textbf{76}          & 79          & 97          & 69          & 81          & 83          & 90          & 95          & \textbf{59}          & 79          & \textbf{84}           \\ 

      &MILe &\textbf{99}&95&\textbf{74}&\textbf{77}&\textbf{94}&64&75&\textbf{69}&\textbf{77}&74&\textbf{87}&94&\textbf{74}&\textbf{83}&\textbf{84}&\textbf{94}&93&56&77&81\\
      \bottomrule
    \end{tabular}
  }

  \caption{\small Mean per-class balanced accuracy in percentage points for each of the
    40 face attributes on CelebA.}
  \label{table:att_acc_pred}
\end{table*}
\section{Comparisons with Noisy Student Scehduling}
\label{app:noisy_student}
\begin{figure}[t]
    \centering
    \includegraphics[width=0.7\linewidth]{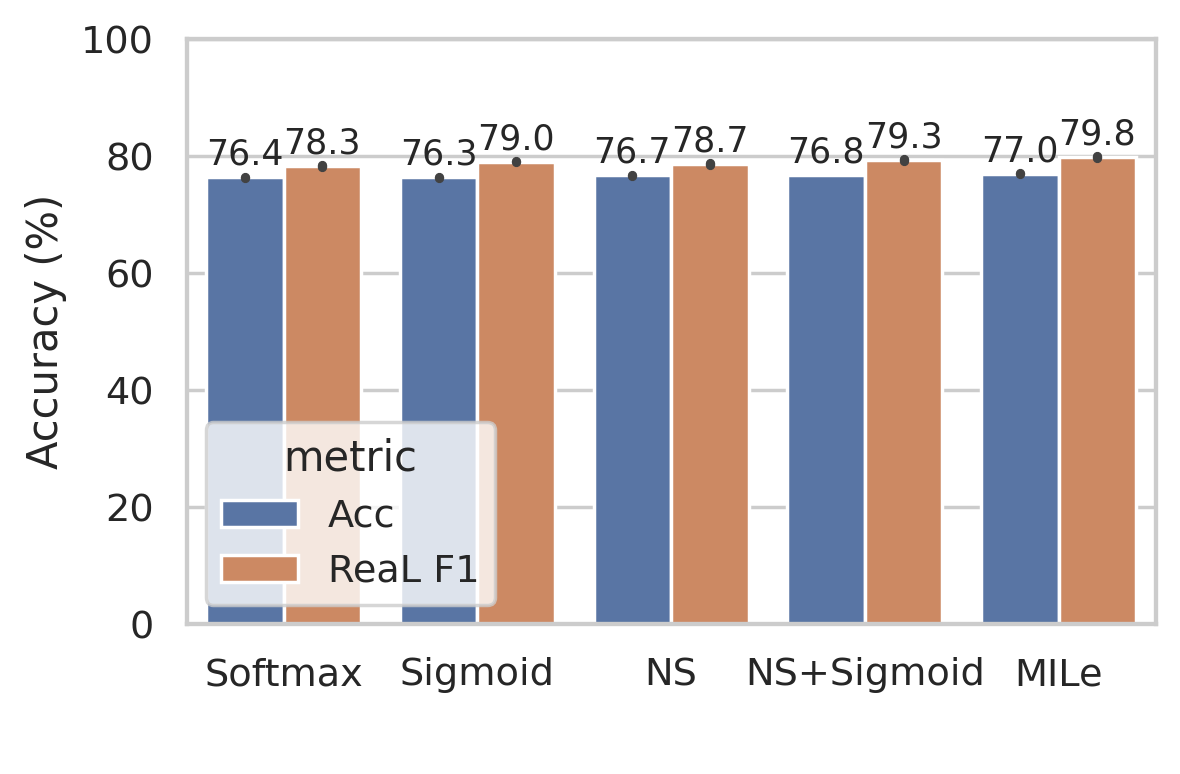}

    \caption{\textbf{Ablation study.} Comparison between different iteration schedules. (a) Comparison with noisy student (NS). (b)(c) Sweep over length of interactive learning phase $k_t$ and length of imitation phase $k_s$. We report the ReaL-F1 score for 10\% (b) and 100\% (c) data fraction.}
    \label{fig:noisy student}
\end{figure}

\citet{xie2020self} introduced noisy student for labeling unlabeled data during semi-supervised learning. This is different from the goal of MILe, which is to construct a new multi-label representation of the images from single labels. Different from MILe, which trains a succession of short-lived teacher and student models, noisy student trains the model three times until convergence. This raises the question of how would MILe perform if it followed noisy student's iteration schedule instead of the one introduced in the main text.

In Fig.~\ref{fig:noisy student} we compare the performance of the best MILe iteration schedule with the NS schedule. We found that MILe achieves the best performance in terms of the ReaL-F1 score. 
\end{document}